\documentclass[11pt]{article}
\pdfoutput=1
\usepackage{hyperref}

\usepackage[final]{corl_2020} 

\usepackage{url}

\usepackage{microtype}
\usepackage{graphicx}
\usepackage{subcaption}
\usepackage{booktabs} 
\usepackage{amsmath}
\usepackage{cleveref}
\usepackage{amsfonts}
\usepackage{enumitem}
\usepackage{algorithmic}
\usepackage{algorithm}
\usepackage[para]{footmisc}
\usepackage{amsthm}

\newcommand{\KLI} {\mathrm{KL}}

\title{Importance Weighted Policy Learning and Adaptation}

\author{
    Alexandre Galashov \\ \texttt{agalashov@google.com} \And Jakub Sygnowski \\ \texttt{sygi@google.com} \And Guillaume Desjardins \\ \texttt{gdesjardins@google.com} \And Jan Humplik \\ \texttt{jhumplik@google.com} \And Leonard Hasenclever \\ \texttt{leonardh@google.com} \And Rae Jeong \\ \texttt{raejeong@google.com} \And Yee Whye Teh \\ \texttt{ywteh@google.com} \And Nicolas Heess \\ \texttt{heess@google.com}
}

\begin{document}
\maketitle


\begin{abstract}
The ability to exploit prior experience to solve novel problems rapidly is a hallmark of biological learning systems and of great practical importance for artificial ones. In the meta reinforcement learning literature much recent work has focused on the problem of optimizing the learning process itself. In this paper we study a complementary approach which is conceptually simple, general, modular and built on top of recent improvements in off-policy learning. The framework is inspired by ideas from the probabilistic inference literature and 
combines robust off-policy learning with a behavior prior, or default behavior that constrains the space of solutions and serves as a bias for exploration; as well as a representation for the value function, both of which are easily learned from a number of training tasks in a multi-task scenario. Our approach achieves competitive adaptation performance on hold-out tasks compared to meta reinforcement learning baselines and can scale to complex sparse-reward scenarios.

\end{abstract}



\section{Introduction}
	
    Current reinforcement learning (RL) algorithms have achieved impressive results across a broad range of games and continuous control platforms.
While effective, such algorithms all too often require millions of environment interactions to learn, requiring access to large compute as well as simulators or large amounts of demonstrations. This stands in stark contrast to the efficiency of biological learning systems \citep{lake_2017}, as well as the need for data-efficiency in real world systems, e.g.\ in robotics where environment interactions can be expensive and risky. In recent years, data efficient RL has thus become a key area of research and stands as one of the bottlenecks for RL to be applied in the real world \citep{dulacarnold2019challenges}. Research in the area is multi-faceted and encompasses multiple overlapping directions. Recent developments in off-policy and model-based RL have dramatically improved stability and data-efficiency of RL algorithms which learn \emph{tabula rasa} \citep[e.g.][]{abdolmaleki2018relative,haarnoja2018sac}. A rapidly growing body of literature, under broad headings such as \emph{transfer learning}, \emph{meta learning}, or \emph{hierarchical RL}, aims to speed up learning by reusing knowledge acquired in previous instances of similar learning problems. \emph{Transfer learning} typically follows a two step procedure:  a system is first \emph{pre-trained} on one or multiple training tasks, then a second step \emph{adapts} the system on a downstream task. While transfer learning approaches allow significant flexibility in system design, the two-step process is often criticised for being sub-optimal. 
In contrast, \emph{meta-learning} incorporates adaptation into the learning process itself. In gradient-based approaches, systems are explicitly trained such that they perform well on a downstream task after a few gradient descent steps~\citep{finn2017maml}. Alternatively, in encoder-based approaches a mapping is learned from a data collected in a downstream task to a task representation~\citep[e.g][]{duan2016rl2,wang2016learning,rakelly2019efficient,zintgraf2019varibad,ortega2019meta,humplik2019meta}.
Because meta-learning approaches optimize the adaptation process directly, they are expected to adapt faster to downstream tasks than transfer learning approaches. But performing this optimization can  be algorithmically or computationally challenging, making it difficult to scale to complex and broader task distributions, especially since many approaches simultaneously solve not just the meta-learning but also a challenging multi-task learning problem.

Given the limitations of meta-learning, a number of recent works have raised the question whether transfer learning methods, potentially combined with data-efficient off-policy algorithms, are sufficient to achieve effective generalization as well as rapid adaptation to new tasks. For example, in the context of supervised meta learning, \citet{raghu2019rapid} showed that learning good features and finetuning during adaptation led to results competitive with MAML. In reinforcement learning, \citet{fakoor2020metaqlearning} showed that direct application of TD3 \citep{fujimoto2018addressing} to maximize a multi-task objective along with a recurrent context and smart reuse of training data was sufficient to match performance of SOTA meta-learning methods on current benchmarks.

In this paper, we take a similar perspective and try to understand the extent to which fast adaptation can be achieved using a simple transfer framework, with the generality of gradient-based adaptation. Central to our approach is the behaviour prior recovered by multi-task KL-regularized objectives \citep{teh2017distral, galashov2018information}. We improve transfer performance by leveraging this prior in two important ways: first, as a regularizer which helps with exploration and restricts the space of solutions that need to be considered, and second as a proposal distribution for importance weighting, where the weights are learnt and given by the exponentiated Q-function. This avoids the need to learn an explicit parametric policy for the transfer task, instead the policy is obtained directly by tilting the prior with the learned, exponentiated action-value function.
To further speed-up adaptation and avoid learning this Q-function de-novo, we make use of a particular parameterization of the action-value functions obtained during multi-task training: the Q-values are parameterized to be linear in some shared underlying feature space. Intuitively, this shared feature representation captures the commonalities in terms of both reward and transition dynamics. In practice, we found this value function representation together with the behaviour prior to generalize well to transfer tasks, drastically speeding-up the adaptation process. We show that across continuous control environments ranging from standard meta-RL benchmarks to more challenging environments with higher dimensional action spaces and sparse rewards, our method can match or outperform recent meta-learning approaches, echoing recent observations in \cite{fakoor2020metaqlearning}.

Our paper is structured as follows. Section~\ref{sec:problem_statement} provides the necessary background material and characterizes the multi-task reinforcement learning problem. Our method, based on importance weighting, is presented in Section~\ref{sec:method} while Section~\ref{sec:transfer} shows how our training algorithm can be adapted to improve transfer learning performance. Relevant work is discussed in Section~\ref{sec:related_work} with experimental results presented in Section~\ref{sec:experiments}.

\section{Background}
\label{sec:problem_statement}

We consider a multi-task reinforcement learning setup,
where we denote a probability distribution over tasks as $\mathcal{P}(\mathcal{T})$. Each task $\mathcal{T} \sim \mathcal{P}$ is a Markov Decision Process (MDP), i.e. a tuple $\left\langle p_{\mathcal{T}}(s' | s, a), p_{\mathcal{T}}(s_{0}), r_{\mathcal{T}}(s, a), \mathcal{A}, \mathcal{S} \right\rangle $ described by (respectively) the transition probability, initial state distribution, reward function, action and state spaces, where $\mathcal{A}$ and $\mathcal{S}$ are identical across tasks. Furthermore, we assume that we are given finite i.i.d. samples of tasks split into training, $\mathcal{T}_{train} = (\mathcal{T}_{1},\ldots,\mathcal{T}_{n})$, and test, $\mathcal{T}_{test} = (\mathcal{T}_{n+1},\ldots,\mathcal{T}_{n+m})$ sets. For each task, denoted by $i$, we denote the task-specific policy as $\pi_{i}$, whereas $\pi_{0}$ is a shared \emph{behaviour prior} which regularizes the $\pi_{i}$'s. On top of that, we denote as $p_{i}(s' | s, a), p_{i}(s_{0}), r_{i}(s, a)$, the transition probability, initial state distribution and reward function for the task $i$.

The starting point in this paper is DISTRAL~\citep{teh2017distral} which aims to optimize the following multi-task objective on the training set:
\begin{equation}
    \label{eq:multitask-kl-objective}
    \mathcal{J}(\pi_{0}, \pi_{1},\ldots,\pi_{n}) = \sum_{i=1}^n \mathbb{E}_{\tau \sim \pi_{i}(\tau)} \left[
    \sum_{t\ge 1} \gamma^t r_i(a_t, s_t) - \gamma^t \alpha \log \frac{\pi_{i}(a_t\mid s_t)}{\pi_{0}(a_t \mid s_t)} \right],
\end{equation}
where $\alpha$ is an inverse temperature parameter and $\tau \sim \pi_{i}(\tau)$ denotes the sampling a trajectory from the task $i$ using the policy $\pi_{i}$. The objective in \eqref{eq:multitask-kl-objective} is optimized with respect to all $\pi_{i}$ and $\pi_{0}$ jointly.
In particular, for each task $i$ and for a fixed behaviour prior $\pi_{0}$, the optimization of the objective $\mathcal{J}$ is equivalent to solving a regularized RL problem with augmented reward  $\tilde{r_{i}}(a_{t}, s_{t}; \pi_{i}) = r_{i}(a_{t}, s_{t}) - \alpha \log \frac{\pi_{i}(a_t\mid s_t)}{\pi_{0}(a_t \mid s_t)}$. As for learning the behaviour prior $\pi_{0}$, optimizing \eqref{eq:multitask-kl-objective} with respect to $\pi_0$ amounts to minimizing the sum of KL divergences between the task-specific policies $\pi_{i}$ and the prior:

\begin{equation}
    \pi^{*}_{0}(a_{t} | s_{t}) = \arg\min_{\pi_{0}} \sum_{i} \KLI[\pi_{i}(a_{t} | s_{t}) || \pi_{0}(a_{t} | s_{t})].
\end{equation}

The behaviour prior's role is to model behavior that is shared across the tasks.
As shown in \citep{galashov2018information}, a prior trained according to \eqref{eq:multitask-kl-objective} with computational restrictions such as partial access to observations only
(information asymmetry) can capture useful default behaviours (such as walking in some walking-related task). The prior regularizes the task-specific solutions and can transfer useful behavior between tasks, which can speed up learning.

Let $\pi_{i}$ be the current policy for the task $i$. For a fixed behaviour prior $\pi_{0}$, we define the associated soft Q-function as 
\begin{equation}
    \label{eq:soft_q_function}
    Q^{\pi_{i}}_{i}(s,a) = r(s,a) + \gamma \mathbb{E}_{s' \sim p_{i}(s' | s,a)} \left[ \mathbb{E}_{a' \sim \pi_{i}(a'|s')}[Q^{\pi_{i}}_{i}(s',a')] - \alpha \KLI[\pi_{i}( \cdot | s') || \pi_{0}( \cdot | s')] \right].
\end{equation}

This function was considered in \citep{fox2015taming}. Note that if $\pi_{0}$ is a uniform distribution, the definition in \eqref{eq:soft_q_function} is equivalent to the soft Q-function considered, for instance, in \citep{haarnoja2018sac,hausman2018learning}. Furthermore, the policy, which is a result of computing 1-step soft-greedy policy, defined as:
\begin{equation}
    \label{eq:improved_policy}
     q(a | s) = \frac{\pi_{0}(a | s) \exp(Q^{\pi_{i}}_{i}(s,a) / \alpha)}{\int \pi_{0}(a | s) \exp(Q^{\pi_{i}}_{i}(s,a) / \alpha) da},
\end{equation}
will have higher soft Q-value on the task $i$, i.e. $Q^{q}_{i}(s,a) \geq Q^{\pi_{i}}_{i}(s,a), \forall a,s$ (see \citep{haarnoja2018sac}). Therefore, \eqref{eq:improved_policy} gives us a principled way to perform policy improvement. 
A similar policy improvement step is used, for instance, in  MPO~\citep{abdolmaleki2018relative} and Soft Actor Critic (SAC)~\citep{haarnoja2018sac}. In both cases, the authors optimize a parametric representation to fit the distribution in \eqref{eq:improved_policy}. 

But instead of fitting a parametric policy, one can directly act according to the improved policy in \eqref{eq:improved_policy}. This can be potentially more efficient, since it avoids an additional step of learning policy with function approximation. However, sampling exactly from the distribution in \eqref{eq:improved_policy} can only be done in a few special cases. Below, we propose a method which uses importance sampling to draw samples from a distribution, which approximates the distribution in \eqref{eq:improved_policy}.

\section{Importance weighted policy learning}
\label{sec:method}
For each task $i$ and for a fixed behaviour prior $\pi_{0}$, we consider the following. Firstly, we sample a set of actions from the behaviour prior:
\begin{equation}
    \label{eq:behaviour_prior_sample}
    \{a^{(k)}\}_{k=1}^K \sim_{iid} \pi_0(a\mid s)
\end{equation}

We denote as $\mathcal{A}_{K} = \{a^{(k)}\}_{k=1}^K$, the set of sampled actions and as $\Pi_{K}(s)$ the set of discrete action distributions defined on $\mathcal{A}_{K}$ for a state $s$. For simplicity of notation, we will drop $s$ from $\Pi_{K}(s)$ and denote it as $\Pi_{K}$. We denote as $Q_{i}$ the soft action value function for some policy $\pi_{i}$ and reward function $r_{i}$. Then, we construct the following action distribution over $\mathcal{A}_{K}$ for each state $s$:
\begin{align}
\label{eq:iwpl}
\hat{q}_{k} = \hat{q}(a = a^{(k)}\mid s) & = \exp\left(\frac{Q_{i}(s, a^{(k)}) -  Z(s) }{\alpha}\right) \text{ for $k=1,\ldots,K$}, \\
&\hat{a} \sim \hat{q}(a|s) = Cat(q_{1},\ldots,q_{K}) \nonumber
\end{align}
with a normalizing constant $Z(s)$:
\begin{equation*}
    Z(s) = \alpha \log \sum_{j=1}^{K} \exp(\frac{Q_{i}(s, a^{(j)})}{\alpha})
\end{equation*}
Then, the resulting policy $\hat{q} \in \Pi_{K}$ is a discrete approximation for the improved policy of the form $q$ from \eqref{eq:improved_policy}. Note that the procedure \ref{eq:iwpl} corresponds to a soft-max distribution over actions with respect to the exponent of the soft Q-function.

In the limit of $K\to\infty$, the procedure \ref{eq:behaviour_prior_sample}-\ref{eq:iwpl} is guaranteed to sample from the policy $q$ from \eqref{eq:improved_policy}. The above sampling scheme gives rise to the \emph{Importance Weighted Policy Learning} (IWPL) algorithm, which combines non-parametric policy evaluation and improvements steps, described below.

\textbf{Non-parametric policy evaluation}
Let $Q: \mathcal{S} \times \mathcal{A} \rightarrow \mathbb{R}$ be a function and $\pi$ is a policy defined on $\mathcal{A}$. We define the soft Bellman backup operator:
\begin{multline*}
    \mathcal{T}^{\pi} Q(s_{t}, a_{t}) = \\ r(s_{t}, a_{t}) + \gamma \mathbb{E}_{s_{t+1}} \left[ \mathbb{E}_{a_{t+1} \sim \pi(\cdot | s_{t+1})} [ Q(s_{t+1}, a_{t+1})]- \alpha KL[\pi(\cdot | s_{t+1}) || \pi_{0}(\cdot | s_{t+1})] \right].
\end{multline*}

It is easy to see (as in \cite{haarnoja2018sac}) that the Bellman iteration $Q^{l+1} = \mathcal{T}^{\pi} Q^{l} ,l\to\infty$ converges to the soft value function \ref{eq:soft_q_function} for $\pi$. Then, for the policy $q$ defined by eq.\ref{eq:improved_policy} we consider an estimator for the Bellman operator induced by the importance weighting procedure \ref{eq:behaviour_prior_sample}-\ref{eq:iwpl} (with a new sampled set of actions $\{a^{(k)}\}_{k=1}^K$):

\begin{equation}
    \label{eq:td_target}
    \mathcal{T}^{q}_{K} Q(s_{t}, a_{t}) = r(s_{t}, a_{t}) + \gamma \mathbb{E}_{s_{t+1}} \left[ \sum_{k=1}^{K} \hat{q}(a_{k} | s_{t+1}) \left( Q(s_{t+1}, a_{k}) - \alpha \log \frac{\hat{q}(a^{(k)} | s_{t+1})}{\pi_{0}(a^{(k)} | s_{t+1})} \right) \right].
\end{equation}

In the limit, this procedure would converge to the soft Q-function for $q$: $Q^{l+1} = \mathcal{T}_{K}^{\pi} Q^{l} ,l\to\infty,K\to\infty$.

\paragraph{Non-parametric policy improvement}
Given the current proposal $\pi_{0}$, some old policy $q^{old}$, corresponding soft Q-function $Q^{q_{old}}$, we can obtain new policy $q^{new}$ via \eqref{eq:improved_policy}. In this case, similar to \citep{hausman2018learning} (Appendix B.2), we have:
\begin{equation*}
    Q^{q_{new}}(s,a) \geq Q^{q_{old}}(s,a), \forall s,a,
\end{equation*}
where $Q^{q_{new}}$ is the soft Q-function corresponding to the $q^{new}$. To approximate the $q^{new}$, we resample new actions $\{a^{(k)}\}_{k=1}^K$ via procedure \ref{eq:behaviour_prior_sample} and apply procedure \ref{eq:iwpl} to the $Q^{q_{old}}$ and obtain the categorical distribution with following probabilities:
\begin{equation*}
    \hat{q}^{new}_{k} = \hat{q}^{new}_{k}(a = a^{(k)} | s)  \sim \exp\left(\frac{Q^{q_{old}}(s; a^{(k)})}{\alpha}\right)
\end{equation*}

This describes a policy improvement procedure based on importance sampling.

\paragraph{Behaviour prior (proposal) improvement}
Given current policy $q(a | s)$ of a form \ref{eq:improved_policy}, corresponding approximation $\hat{q}$ from \eqref{eq:iwpl}, a new behaviour prior $\hat{\pi}_{0}$ is obtained by maximizing the likelihood of obtaining samples from $\hat{q}(a|s)$:
\begin{equation*}
    \hat{\pi}_{0}(\cdot | s) = \arg\min_{\pi_{0}}\sum_{k=1}^{K} \hat{q}_{k} \log \pi_{0}(a_{k} | s)
\end{equation*}

\paragraph{Temperature calibration}

In the current formulation, IWPL requires us to choose the inverse temperature parameter in \ref{eq:multitask-kl-objective} and in \ref{eq:iwpl}. For varying reward scales, it could result in an unstable behaviour of the procedure \ref{eq:iwpl}.
Some RL algorithms, such as REPS~\citep{reps2010}, MPO~\citep{abdolmaleki2018relative} 
therefore replace similar (soft) regularization terms with hard limits on KL or entropy. Here, we consider a hard-constraint version of objective (\ref{eq:multitask-kl-objective}): 

\begin{align}
    \label{eq:hard_kl_objective}
    & & \sum_{i} \mathbb{E}_{\tau \sim \pi_{i}(\tau)} \left[\sum_{t\ge 1} \gamma^t r_i(a_t, s_t) \right] & & \\
    & & \sum_{i} \mathbb{E}_{s \sim \pi_{i}(s)} \KLI[\pi_{i}(\cdot | s) || \pi_{0}(\cdot | s)] < \epsilon & & \nonumber
\end{align}

The parameter $\epsilon$ defines the maximum average deviation of all the policies $\pi_{i}$ from the behaviour prior $\pi_{0}$. Given $\epsilon$, we can adjust the inverse temperature $\alpha$ to match this constraint. In many cases $\epsilon$ is easier to choose than the inverse temperature $\alpha$ since it does not, for instance, depend on the scale of the reward. The associated temperature parameter $\alpha$ can be optimized by considering the Lagrangian for the objective \ref{eq:hard_kl_objective}, similar to REPS~\citep{reps2010} and MPO~\citep{abdolmaleki2018relative}.

\paragraph{Algorithm} The concrete algorithm is a combination of the steps above with parametric function approximation of the necessary quantities. We consider $\pi_{0}(a|s, \phi)$ the approximation for the behaviour prior $\pi_{0}$ and $Q_{\theta_{i}(s,a)}$ an approximation for the soft value function for the task $i$. We denote as $\phi'$ and as $\theta_{i}'$ the other set of parameters which correspond to the target networks (see \citet{mnih2013playing}) - the networks which are kept fixed for some number of iterations. We denote as $\hat{q}_{i}'$ the discrete policy coming from \ref{eq:iwpl} associated with $Q_{\theta_{i}'}(s,a)$ and $\pi_{0}(a|s, \phi')$. Then, $Q_{\theta_{i}}(s,a)$ can be trained by minimizing the Bellman residual:
\begin{equation}
    \label{eq:critic_loss}
     \mathcal{J}_{Q}(\theta) = \sum_{i} \mathbb{E}_{s,a \sim p_{i}(s,a)} \left[ \frac{1}{2} (Q_{\theta_{i}}(s,a) - \hat{Q}_{i}(s,a))^{2} \right],
\end{equation}
where $\theta=(\theta_{1},\ldots,\theta_{n})$ and:
\begin{equation}
    \hat{Q}_{i}(s,a) = r_{i}(s,a) +\gamma \sum_{k=1}^{K} \hat{q}_{i}'(a^{(k)}_{i} | s_{t+1}) \left(  Q_{\theta_{i}'}(s_{t+1}, a^{(k)}_{i}) - \alpha \log \frac{\hat{q}_{i}'( a^{(k)}_{i} | s_{t+1})}{\pi_{0}( a^{(k)}_{i} | s_{t+1}, \phi')} \right)
\end{equation}
The behaviour prior $\pi_{0}(a | s, \phi)$ is learned by minimizing:
\begin{equation}
    \mathcal{J}_{\pi_{0}}(\phi) = - \sum_{i} \mathbb{E}_{s \sim p_{i}(s)} \left[  \sum_{k=1}^{K} \hat{q}'_{i}(a_{k} | s) \log \pi_{0}(a_{k} | s, \phi) \right]
\end{equation}

The full algorithm is presented in Algorithm~\ref{alg:iwpl}.

\begin{algorithm}[tb]
\caption{Distributed Importance Weighted Policy Learning (IWPL)}
\label{alg:iwpl}
\begin{algorithmic}
    \STATE {\bfseries Input:}
    \STATE Behaviour prior $\pi_{0}(a | s, \phi)$, initial parameters $\phi_{0}$
    \STATE Q-function $Q_{\theta_{i}}$, initial parameters $\theta^{0}_{i}$ for each task $i$
    \STATE Target networks with a separate set of parameters $\theta'$, $\phi'$
    \STATE Target networks update period $T$
    \STATE Learning rates $\beta_{Q}, \beta_{\pi_{0}}, \beta_{\alpha}$
    \STATE Replay buffer $\mathcal{B}$ containing data $\mathcal{B}_{i}$ for each task $i$
    \STATE Training tasks indexes $\mathcal{I}=\{1,\ldots,n\}$
    \STATE Define $\theta=(\theta_{1},\ldots,\theta_{n})$, $\theta'=(\theta'_{1},\ldots,\theta'_{n})$
    \STATE {\bfseries Steps:}
    \STATE \textbf{Actor policy}:
    \WHILE{Not converged}
        \STATE Receive parameters from the learner
        \STATE Sample uniformly a training task $i$ from $\mathcal{I}$
        \STATE Sample full-episode trajectory $\tau=(s_{0},a_{0},r_{0},\ldots,s_{T},a_{T},r_{T}) \sim \hat{q}_{i}(\tau)$, using equations.~(\ref{eq:behaviour_prior_sample},\ref{eq:iwpl})
        \STATE $\mathcal{B}_{i} = \mathcal{B}_{i} \cup \tau$
    \ENDWHILE
    \STATE \textbf{Learner policy}:
    \WHILE{Learning}
        \STATE Sample uniformly (with replacement) a batch of tasks $\mathcal{I}_{b}$ from  $\mathcal{I}$
        \FOR{each task $i$ from $\mathcal{I}_{b}$}
        \STATE Sample partial trajectory from replay buffer $\mathcal{B}_{i}: \tau_{t:t+M} = (s_{t}, a_{t}, r_{t},\ldots, r_{t+M})$ for task $i$
        \STATE Sample $K$ actions $(a^{t}_{1},\ldots,a^{t}_{K})$ from $\pi_{0}(a | s_{t}, \phi')$, for each state $s_{t}$
        \STATE Calculate the $Q_{\theta'_{i}}(s_{t}, a^{t}_{k}), \forall t,k$
        \STATE Construct categorical distribution $\hat{q}_{i}'$ as in \eqref{eq:iwpl} using $ Q_{\theta'}(s_{t}, a^{t}_{k})$
        \STATE \emph{\% Perform gradient update on the parameters}
        \STATE $\theta_{i} \leftarrow \theta_{i} + \beta_{Q} \nabla_{\theta_{i}}
        \mathcal{J}_{Q}(\theta)$
        \STATE $\phi \leftarrow \phi + \beta_{\pi_{0}} \nabla_{\phi} \mathcal{J}_{\pi_{0}}(\phi)$
        \STATE Every $T$ gradient steps, update target networks parameters $\theta' \leftarrow \theta$, $\phi' \leftarrow \phi$. 
        \ENDFOR
    \ENDWHILE
\end{algorithmic}
\end{algorithm}

\section{Importance weighted policy adaptation for transfer learning}
\label{sec:transfer}

Given pretrained action-value functions  $\{Q_i^\star\}_{i=1}^n$ and a behaviour prior  $\pi_0^\star$ from optimization of the objective \ref{eq:hard_kl_objective} on the training set, we show how to leverage it to quickly solve tasks from the test set. We call this process adaptation. Below, we describe how adaptation is facilitate by two components of our method, behaviour and value transfer.
\paragraph{Behaviour Transfer.}
Given a pre-trained behaviour prior $\pi_0^\star$, we can learn the solution to a new task by learning a new value function and sampling from the implicit policy defined by \ref{eq:iwpl}. This can be achieved by executing the procedure in  Section~\ref{sec:method} without the prior improvement step. Because the policy essentially is initialized from the behaviour prior, the latter constrains possible solutions and leads to sensible exploration. In order to obtain new optimal policy, we need to learn new optimal soft Q function, which can require considerable amount of samples when Q is naively parameterized by a neural network. Below, we propose a way to leverage the Q-functions learned for tasks in the training set to speed up transfer in terms of number of interactions with the environment.
 \paragraph{Value Transfer.} 
 In order to acquire knowledge about the value function that can be leveraged for transfer we choose to represent the task specific value $Q_i$ as a linear function of task-specific parameters $w$ and shared features $\psi$:
\begin{equation}
\label{eq:q_func_parameterisation}
    Q_i(s, a; \Phi_i) = \psi(s,a; \theta)^T w_i,
\end{equation}
where $\psi_\theta: \mathbb{R}^S \times \mathbb{R}^A \rightarrow \mathbb{R}^d$ is a function mapping states and actions to a feature vector (with parameters $\theta$ shared across tasks), $w_i \in \mathbb{R}^d$ is a task-specific vector used to identify task-specific Q-values, and $\Phi_i=\{\theta, w_i\}$. During the adaptation phase, we initialize $Q(s,a)$ as $\psi(s,a;\theta^\star)^\top \tilde{w}$, with $\tilde{w} \sim \mathcal{N}(0, I_d/d)$, and adapt $\tilde{w}$ using TD(0) learning. Furthermore, for some more challenging tasks, we replace (at training time) the task-specific vector $w_{i}$ by a non-linear embedding of a structured goal descriptor $g_i$ which is available during training but not during adaptation, i.e. $Q_i(s, a, g_i; \Phi_i) = \psi(s,a; \theta)^\top f(g_i; \theta)$, where $f(g_i; \theta)$ is a learned embedding of goal $g_i$ with parameters $\theta$ shared across training tasks. At test time, we initialize the critic as before: $\psi(s,a;\theta^\star)^\top \tilde{w}$. Since some RL problems can still be challenging multi-task learning problems, this "asymmetry" between learning and testing allows us to simplify the solution of the multi-task problem without affecting the applicability of the learned representation,
in contrast to most of the meta-learning approaches which require that training and adaptation phase be matched. Then, our proposed method exploits both, behaviour prior and shared value features to derive an efficient off-policy transfer learning algorithm. Note that this approach does not require to have a finite or/and discrete set of tasks and could work also in the continuously parameterised task distributions, since we essentially allow the task-specific Q-function to depend on the task conditioning.

\paragraph{Algorithm}
Given the new task $j$, we will learn associated $w$ to construct Q-function of the form \ref{eq:q_func_parameterisation}. Let $\pi_{0}(a|s, \phi)$ be a pretrained behaviour prior, $\psi(s,a; \theta)$ be pretrained features for the Q-functions on the training set. We use similar notation as in Section~\ref{sec:method}, by denoting as $w'$, the target network parameters and as $\hat{q},\hat{q}'$ associated categorical distributions of form \ref{eq:iwpl}. Let $Q_{w}(s,a; \theta)$ be the function approximator of the form \ref{eq:q_func_parameterisation} for the new task $j$. Then, the adaptation on the task $j$ reduces to learning the Q-function by minimizing TD(0) Bellman residual:
\begin{equation}
    \label{eq:adaptation_critic_loss}
     \mathcal{J}(w; \theta) = \mathbb{E}_{s,a \sim p(s,a)} \left[ \frac{1}{2} (Q_{w}(s,a; \theta) - \hat{Q}_{w'}(s,a; \theta'))^{2} \right],
\end{equation}
where
\begin{equation}
    \hat{Q}_{w'}(s,a; \theta') = r_{j}(s,a) +\gamma   \sum_{k=1}^{K} \hat{q}_{j}'(a^{(k)}_{j} | s_{t+1}) \left( Q_{w'}(s_{t+1}, a^{(k)}_{j}; \theta') - \alpha \log \frac{\hat{q}_{j}'(a^{(k)}_{j} | s_{t+1})}{\pi_{0}(a^{(k)}_{j} | s_{t+1}, \phi)} \right).
\end{equation}

Note that in addition to learning new $w$, it is also possible to finetune pre-trained features $\psi(s,a; \theta)$. It may be required if test tasks are too different from the training tasks. This scenario is discussed in \emph{Generalization} part of Section~\ref{sec:experiments}. We call the resulted algorithm \emph{Importance Weighted Policy Adaptation} (IWPA) which is described in Algorithm~\ref{alg:iwpa}.

\begin{algorithm}[tb]
\caption{Importance Weighted Policy Adaptation (IWPA)}
\label{alg:iwpa}
\begin{algorithmic}
    \STATE {\bfseries Input:}
    \STATE  Behaviour prior $\pi_{0}(a|s; \phi)$ pre-trained on the training set.
    \STATE  Shared features $\psi(s, a; \theta)$ representing optimal training soft Q-functions \ref{eq:q_func_parameterisation}
    \STATE $\mathcal{I} = \{n+1,\ldots,n+m\}$ - indexes for the test set tasks.
    \STATE $N$: Number of adaptation episodes
    \STATE $M$: Number of gradient updates
    \STATE Target networks parameters $w', \theta'$
    \STATE Target networks update period $T$
    \STATE $\beta_{w}, \beta_{\theta}$ - Learning rates
    \FOR{Each test task $j$ from $\mathcal{I}$}
        \STATE Initialize task specific critic parameters $w \sim \mathcal{N}(0, I_d/d)$
        \STATE Define action-value function $Q_{w}(s,a; \theta) = \psi(s, a; \theta)^T w$
        \STATE Denote as $\hat{q}_{w}$ associated to $\pi_{0}$ and $Q_{w}$ categorical distribution of form \ref{eq:iwpl}
        \FOR{$n=1:N$}
            \STATE Sample full-episode trajectory $\tau=(s_{0},a_{0},r_{0},\ldots,s_{T},a_{T},r_{T}) \sim \hat{q}_{w}(\tau)$, using eqs.~(\ref{eq:behaviour_prior_sample},\ref{eq:iwpl})
            \FOR{$m=1:M$}
                \STATE \emph{\% Perform gradient update on the parameters for adaptation}
                \STATE $w \leftarrow w + \beta_{w} \nabla_{w} \mathcal{J}(w, \theta)$
                \STATE (Optionally) Finetune features, $\theta \leftarrow \theta + \beta_{\theta} \nabla_{\theta} \mathcal{J}(w, \theta)$
                \STATE Every $T$ gradient steps, update target networks parameters $w' \leftarrow w$, $\theta' \leftarrow \theta$. 
            \ENDFOR
        \ENDFOR
   \ENDFOR
\end{algorithmic}
\end{algorithm}

\section{Related Work}
\label{sec:related_work}

The proposed algorithm has some similarities to recent off-policy RL methods. In both Maximum a Posteriori Policy Optimization (MPO)~\citep{abdolmaleki2018relative} and in Soft Actor Critic (SAC)~\citep{haarnoja2018sac}, the authors propose to learn the parametric policy and fit it to the non-parametric improved policy as in eq.~ \ref{eq:improved_policy} (in MPO, the $\pi_{0}$ is replaced by the parametric policy, whereas in SAC, $\pi_{0}$ is replaced by the uniform distribution). Furthermore, as in our method, in SAC the authors use induced soft Q-function. The both methods collect the experience using the parametric policy. In contrast, in our method, we directly use the improved non-parametric policy to collect the experience as well as to construct the bootstrapped Q-function. Moreover, our method is explicitly build in the context of multi-task learning and makes use of behaviour prior with information asymmetry~\citep{galashov2018information} which encourages structured exploration.

In recent work on Q-learning, there were many attempts to scale it up to high-dimensional and continuous action domains.
In soft Q-learning~\citep{haarnoja2017reinforcement}, in the context of maximum entropy RL, the authors learn a parametric mapping from normally-distributed samples to ones drawn from a policy distribution, which converges to the optimal non-parametric policy induced by a soft Q function (in a similar way as in eq. \ref{eq:improved_policy} with a uniform $\pi_{0}$). In Amortized Q-learning \cite{wiele2020qlearning}, the authors propose to learn a proposal distribution for actions and then select the one maximizing the Q-function. Unlike in our work, the authors do not regularize the induced non-parametric distribution to stay close to the proposal. Note that, in the limit of the temperature $\tau \rightarrow 0$, then our softmax operator over importance weights becomes a max, making our approach a strict generalization of AQL. Finally, \citet{hunt2018composing}, propose to learn a proposal distribution which is good for transfer to a new task, in the context of successor features~\citep{barreto2017successor} while maximizing the entropy.

Transfer of knowledge from past tasks to future ones is a well-established problem in machine learning \citep{caruana1997multitask, baxter2000model} and has been addressed from several different angles. Meta learning approaches try to learn the adaptation mechanism by explicitly optimizing either for minimal regret during adaptation or for performance after adaptation. Gradient-based approaches, often derived from MAML, aim at learning initial network weights such that a few gradient steps from this initialization is sufficient to adapt to new tasks \citep{finn2017maml, finn2018probabilistic, gupta2018meta, nichol2018reptile}. Memory-based meta learning approaches model the adaptation procedure using recurrent networks \citep{duan2016rl2, wang2016learning, mishra2018simple, humplik2019meta, rakelly2019efficient}. One problem of meta learning approaches is the explicit optimization for adaptation on a new task, which may be computationally expensive. In addition, most of the meta-learning methods require the training and adaptation process to be matched. It could restrict the class of problems which can be solved by this approach since some hard meta RL problems could also constitute hard multi-task problems. Our method allows to provide additional information at training time to facilitate this learning without affecting the adaptation phase.

Other transfer learning methods (ours included) do not explicitly optimize the algorithm for adaptation. A common approach is to use a neural network which shares some parameters across training tasks and fine-tunes the rest. Recent work \citep{raghu2019rapid} suggests that this yields performance comparable to the MAML-style training. Transfer learning with Successor Features \citep{barreto2017successor} exploits a similar decomposition of the action-value function, but relies on Generalized Policy Improvement for efficient transfer, instead of our more general gradient-based adaptation. Another approach for reusing past experience is hierarchical RL which tries to compress the experience to a shared low-level controller or a set of options which are reused in later tasks \citep{brunskill2014pac, heess2016learning, tirumala2019exploiting, wulfmeier2019regularized}. Finally, an approach we build upon is to distill past behavior into a prior policy \citep{teh2017distral, galashov2018information} from which we can bootstrap during adaptation.
In \citet{fakoor2020metaqlearning}, the authors propose a transfer learning approach based on fine-tuning a critic acquired via a multi-task objective. To speed-up adaptation, their method makes heavy use of off-policy data acquired during meta-training, and an adaptive trust region which regularizes the critic parameters based on task similarity.

\section{Experiments}
\label{sec:experiments}

In this section, we empirically study the performance of our method in the following scenarios. Firstly, we assess how well the method performs in the multi-task scenario. Then, we demonstrate the methods ability to achieve competitive performance in adapting to hold-out tasks compared to meta reinforcement learning baselines on a few standard benchmarks. On top of that, we show that the method scales well to more challenging sparse reward scenarios and achieves superior adaptation performance on hold out tasks compared to considered baselines. Finally, we consider the case when the number of training tasks is very small. In this case the behaviour prior and value-function representation may overfit to the training tasks. We demonstrate that our method still generalizes to hold-out task when additional fine-tuning is allowed.
\paragraph{Task setup.}
We consider two standard meta reinforcement learning problems: 2D point mass navigation and half cheetah velocity task, described in \citet{rakelly2019efficient}. In addition to these simple tasks, we design a set of sparse reward tasks, which are harder as control and exploration problems: \emph{Go To Ring}: a quadruped body needs to navigate to a particular (unknown) position on a ring. \emph{Move Box}: a sphere-like robot must move a box to a specific position. \emph{Reach}: a simulated robotic arm is required to reach a particular (unknown) goal position. \emph{GTT}: A humanoid body needs to navigate to a particular (unknown) position on a rectangle. For every task, we consider a set of training $\mathcal{T}_{1},\ldots,\mathcal{T}_{n}$, and held-out tasks $\mathcal{T}_{n+1},\ldots,\mathcal{T}_{n+m}$. For every task, the policy receives proprioceptive information, as well as the global position of the body and the unstructured task identifier (a number from $1$ to $n$). For the \emph{Move Box} task, we provide additional global position of the target as task observation on training distribution to facilitate learning. We do not provide this information when working on test tasks. For more environment details, please refer to Appendix \ref{sec:envs}.
\paragraph{Multi-task training.}
We first demonstrate our method ability to solve multi-task learning problems.
As baseline, we consider SVG(0)~\citep{heess2015learning}, an actor-critic algorithm with additional Retrace off-policy correction~\citep{munos2016retrace} for learning the Q-function as described in \cite{riedmiller2018learning}. We refer to this algorithm as \emph{RS(0)}.
We further consider a continuous-action version of DISTRAL~\citep{teh2017distral} built on top of \emph{RS(0)}, where we learn a behaviour prior alongside the policy and value function, similar to \citep{galashov2018information}. This prior exhibits information asymmetry of observations with respect to the policy and the value function (it receives less information) which makes it to learn useful default behaviour speeding up the learning.
In Appendix~\ref{sec:envs}, we specify the information provided to the behaviour prior and the policy.
Furthermore, we consider MPO~\citep{abdolmaleki2018relative} algorithm as well as its version with behaviour prior, which we call MPO + DISTRAL. The latter simply uses KL-regularizion to the learned prior (alongside the policy learning) in the M-step as soft constraint as well as soft Q-function. In our method, IWPL, we also use the behaviour prior with information asymmetry between Q-function, which receives task-specific information.

For each of the models, we optimize hyperparameters and report the best found configuration with 3 random seeds.
The experiments are run in a distributed setup with 64 actors that generate experience and a single learner somewhat similar to  ~\citet{espeholt2018impala} using. We use a replay buffer of size $10^{6}$ and control the number of times an individual experience tuple is considered by the learner. This ensures soft-synchronicity between ator and learner and ensures a fair comparison between models that differ with respect to the compute cost of inference and learning.
For more details, please refer to the Appendix~\ref{sec:experimental_details}.

\begin{figure}[t!]
    \centering
    \includegraphics[width=0.9\linewidth]{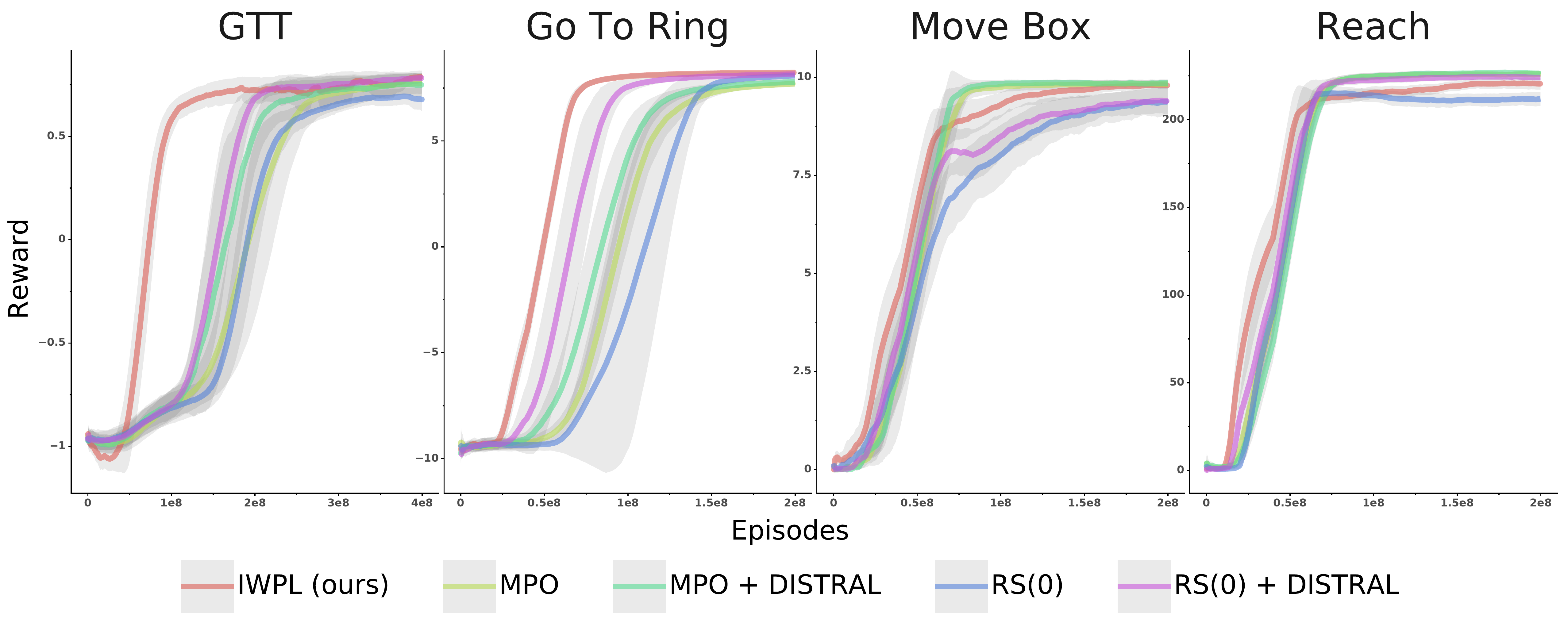}
    \caption{Multi-task training results.} 
    \label{fig:multi_task_training}
\end{figure}

The results are given on Figure~\ref{fig:multi_task_training}. We can see that our method achieves competitive performance compared to the baselines. Note that it has larger gains in tasks where the control problem is harder. This effect of behaviour prior was observed in \citep{galashov2018information} and presumably is amplified for IWPL, where there is no intermediate parametric policy in the loop. It immediately samples the useful actions from the prior which is learned faster than the agent policy due to the restricted set of observations as discussed in \citep{galashov2018information}. Interestingly, we do not observe a difference between MPO and MPO+DISTRAL, presumably because the effect of the behaviour prior is reduced by the hard KL constraint to the previous policy.
\paragraph{Adaptation performance.}
Next, we investigate performance of our method in adapting to hold-out tasks. The main criteria is the data efficiency in terms of a number of episodes on a new task. As discussed in Section \ref{sec:transfer}, we want to leverage the behaviour prior as well as learned shared representation for the action-value function. Therefore, we consider two variants of our method, IWPA described in Section~\ref{sec:transfer}. We refer to "Shared Q + IW" as the version which leverages both behaviour prior and action-value function, and "IW", which leverages only behaviour prior and learns action value function from scratch without making assumption~\ref{eq:q_func_parameterisation}. As natural baseline, we consider RS(0) + DISTRAL agent as in multi-task learning where for learning Q-function we use TD(0) as in IWPA. Starting from this, we call "Shared Q", the agent which leverages both behaviour prior and action-value function and "DISTRAL" which leverages only behaviour prior.

We pre-train "RS(0) + DISTRAL" agent with Q-function parameterisation~\ref{eq:q_func_parameterisation} on the training set, choose best performing hyperparameter and freeze pretrained $\pi_{0}$ and action-value features $\psi$ for each task. Then we apply all four proposed adaptation methods to these behaviour prior and action-value features. The reason to use one algorithm for pretrainining is to isolate the adaptation performance from the multi-task performance studied above. Empirically, we found that models trained based on IWPL lead to similar results, but we decided to report the results pretrained using "RS(0) + DISTRAL" because this agent was already considered in \citep{galashov2018information}.

In addition, we consider two meta-reinforcement learning baselines: a re-implementation of RL2~\citep{duan2016rl2}, \citep{wang2016learning} as well as a re-implementation of PEARL~\citep{rakelly2019efficient}. For both implementations we build upon  RS(0) as the base algorithm. In our implementation of PEARL (denoted as PEARL*), we use simple LSTM to encode the context. As reported in \citet{rakelly2019efficient}, this variant is slower to learn but eventually achieves similar to PEARL performance. Despite this change, our results achieve comparable performance to those presented in Section 6.3 of \citep{rakelly2019efficient}. On top of that, we also consider a baseline which learns to solve the test tasks "From Scratch" and corresponds to  RS(0) algorithm without pre-training and behaviour prior. For more details, see Appendix \ref{sec:experimental_details}.

\begin{figure}[t!]
    \centering
    \includegraphics[width=0.9\linewidth]{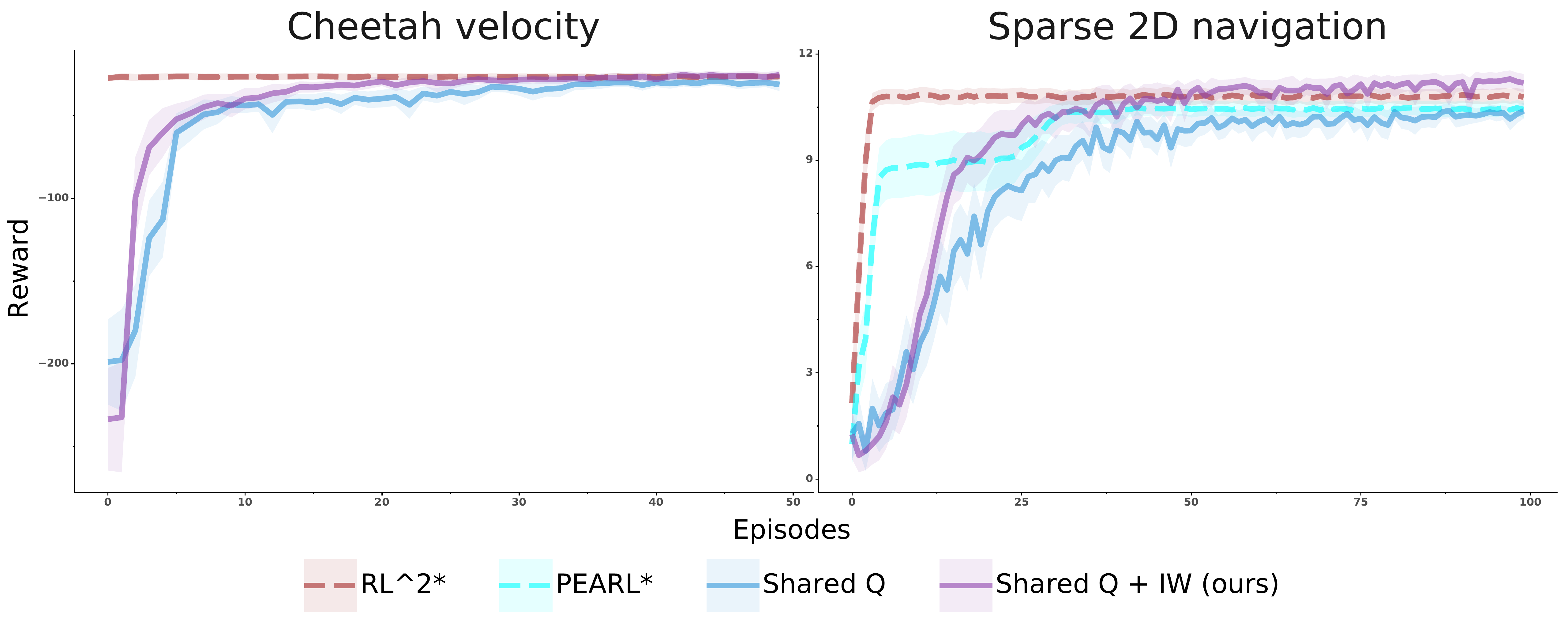}
    \caption{Adaptation performance on standard benchmarks after meta-training. Our method (not using meta-learning) achieves comparable results to other meta-learning baselines.} 
    \label{fig:meta_testing_theirs}
\end{figure}

We start by presenting test-time adaptation performance on two standard continuous control tasks used in \cite{rakelly2019efficient}: \emph{half-cheetah velocity} and \emph{Sparse 2D navigation}. Note, that for \emph{Sparse 2D navigation} task, PEARL receives dense reward during training whereas our agent is trained with sparse rewards. It additionally demonstrates that our method can be employed in more difficult scenarios. The results are presented in Figure \ref{fig:meta_testing_theirs}. While RL2 and PEARL converge faster in absolute terms, IWPA remains competitive and converges quickly despite not optimizing the adaptation process directly.

\begin{figure}[t!]
    \centering
    \includegraphics[width=0.9\linewidth]{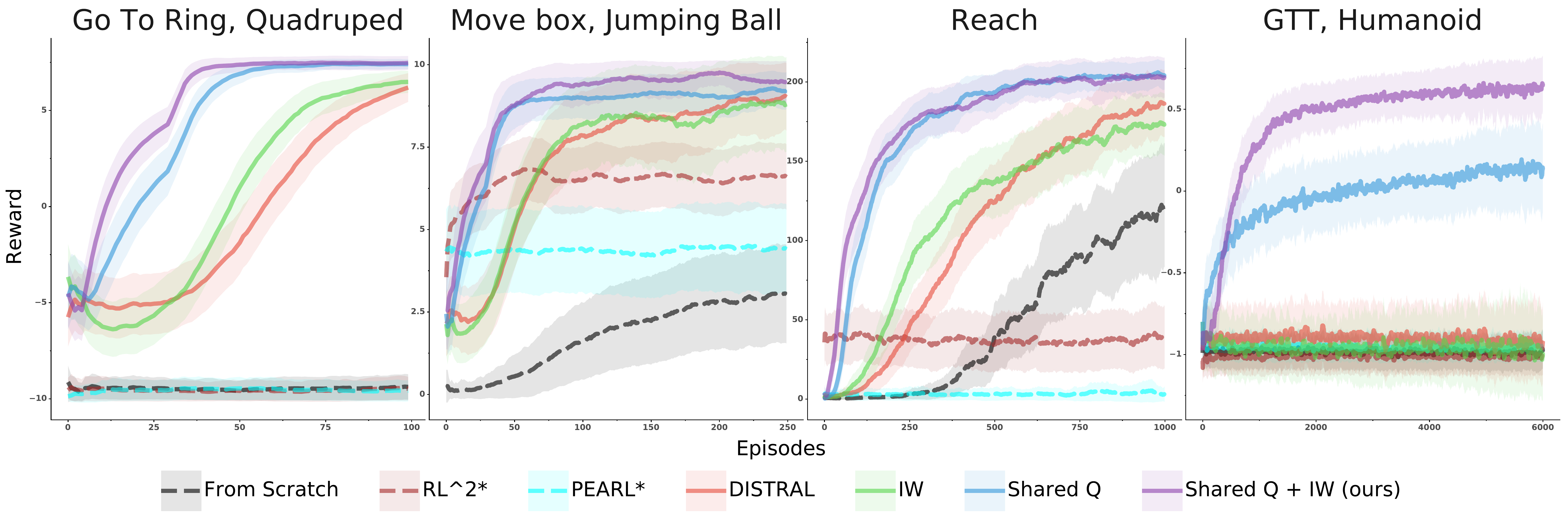}
    \caption{Adaptation performance of different methods on sparse reward tasks after meta-training.} 
    \label{fig:meta_testing_our}
\end{figure}

Going further, we present the results on complex sparse reward tasks. 
Results on these tasks are depicted on Figure \ref{fig:meta_testing_our}. Our proposed method achieves gains in adaptation time with respect to the baseline DISTRAL. Furthermore, we note that using shared features for the value function provides a significant gain. It is important to note that using shared features without the behaviour prior fails to learn fast, because the behaviour prior plays a crucial role in facilitating exploration (see Appendix~\ref{sec:appendix_ablation}). On top of that, we observe that IWPA similarly to multi-task results section, provides bigger gains on harder to control problems, like GTT humanoid. Note that this is a very challenging task: humanoid needs to locate a target and only receives a reward when successfull. Furthermore, the humanoid may fail at any moment and the episodes will terminate. It makes it extremely hard to learn without any prior knowledge. We note that both RL2 and PEARL failed to achieve optimal performance on these tasks. This could be for a variety of reasons, including the sparsity of the rewards and the complexity of learning a single policy that has to operate over long time horizons.

\begin{figure}[t!]
    \centering
    \includegraphics[width=0.9\linewidth]{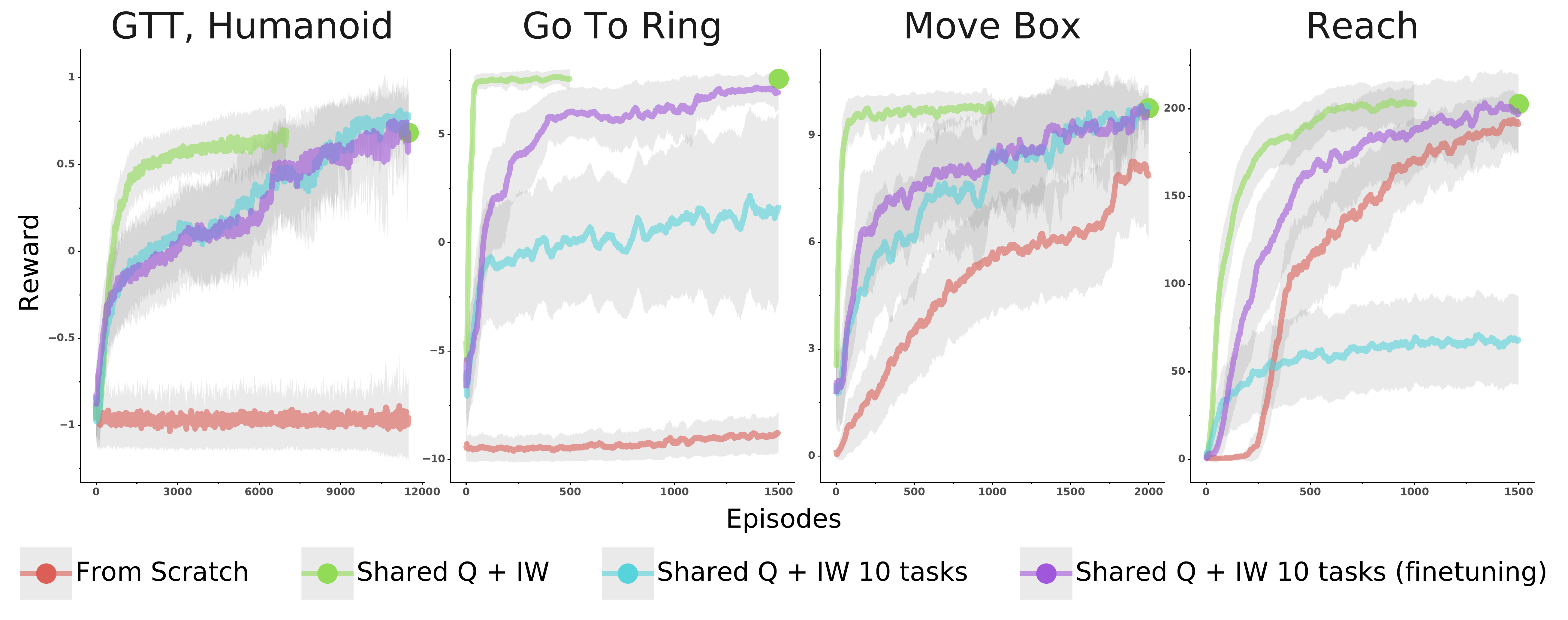}
    \caption{Generalization results. We report the performance of learning from scratch as well as Shared Q + IW architecture trained in high task regime. On top of that, we show the performance of the architectures trained in the low task regime with and without a finetuning of value function features. We denote by a point the final performance of the early-stopped Shared Q + IW experiment.} 
    \label{fig:generalization}
\end{figure}

\paragraph{Generalization}
An efficient transfer learning method should be robust to low data regime. Here we show that in case, when a few of training tasks are available, the method is still be able to generalize if we allow for the additional finetuning of the shared features for the Q-function after 20 episodes of interaction on a new task. For each of the sparse reward tasks, we consider a version which has few training tasks. We trained IWPL on these and compare it to the IWPL trained in large tasks regime. The results are given in Figure \ref{fig:generalization}. As we see, the method trained in a low tasks regime fails to generalize in most of the tasks, whereas the additional finetuning helps to recover the final performance and still be able to do it faster than learning from scratch.

\section{Discussion}
\label{sec:discussion}

We have presented a novel method for multi-task learning as well as for adaptation to new hold-out tasks which does not explicitly meta-learn the adaptation process and yet can  match the adaptation speed of common meta-reinforcement learning algorithms. Instead of explicit meta-learning, we relied on feature reuse and bootstrapping from a behavioral prior. The behavior prior can be seen as an informed proposal for a task distribution that is then specialized to a particular task by a learned action-value function. This scheme can be easily integrated into different actor-critic algorithms for data efficient off-policy learning at training and test time. It further does not strictly require to execute test time adaptation as an inner loop during training thus adding extra flexibility.
\clearpage

\bibliography{main}  

\appendix

\section{Experimental details}
\label{sec:experimental_details}

For all the models, we use similar architectures for all the components. Each agent has actor, critic and optionally behaviour prior networks. For all the methods, except for $RL^2$~\citep{wang2016learning} and PEARL~\citep{rakelly2019efficient}, actor, critic and behaviour prior networks are 2 dimensional multi-layer perceptron with ELU activation followed by one-dimensional linear layer. On top of that, for each of the networks, we use a layer normalizing inputs. For $RL^2$~\citep{wang2016learning}, the actor and critic networks are 2-dimensional multi layer perceptrons with ELU activations, followed by an LSTM with elu activations. In PEARL~\citep{rakelly2019efficient}, actor and critic networks have similar structure as other methods and the encoder network is an LSTM followed by one-dimensional stochastic layer encoding Gaussian distribution. Actor and behaviour prior are represented by Gaussian distributions as well.

\subsection{Multi-task training experiment}

We consider the following hyperparameter ranges:
\begin{itemize}
    \item Learning rates: $1e-3, 1e-4, 5e-4, 5e-5, 1e-5$
    \item Initial inverse temperature $\alpha$: $100, 10, 1, 1e-1, 1e-2, 1e-3, 1e-4$
    \item Epsilon $\epsilon$: $100, 50, 30, 10, 5, 1, 0.1, 0.01, 0.001, 0.0001$
    \item KL-cost (inverse temperature) for DISTRAL baseline $\alpha$: $1, 0.1, 0.01, 0.001, 0.0001, 0.00001, 0.0$
\end{itemize}

For the multi-task experiments, we found that the following values worked best for all the architectures:
\begin{itemize}
    \item Learning rate: $5e-4$
    \item Epsilon $\epsilon$: $50$
\end{itemize}

The best hyperparameters for RS(0) + DISTRAL for multi-task experiment:
\begin{itemize}
    \item Go to Target, Humanoid: $\alpha=0.001$
    \item Go to Ring, Qudruped: $\alpha=0.001$
    \item Move box, Jumping Ball: $\alpha=0.001$
    \item Reach: $\alpha=0.1$
\end{itemize}

The best hyperparameters for IWPL for multi-task experiment:
\begin{itemize}
    \item Go to Target, Humanoid: $\epsilon=100, \alpha=0.1$
    \item Go to Ring, Qudruped: $\epsilon=30, \alpha=0.01$
    \item Move box, Jumping Ball: $\epsilon=1, \alpha=0.01$
    \item Reach:$\epsilon=50, \alpha=0.0001$
\end{itemize}

To have a fair comparison, we optimize E-step epsilon as well as KL cost for MPO~\citep{abdolmaleki2018relative}. We consider the same ranges as above and the best hyperparameters are:
\begin{itemize}
    \item Go to Target, Humanoid:  $\epsilon=0.01, \alpha=0.0001$
    \item Go to Ring, Qudruped: $\epsilon=0.1, \alpha=0.001$
    \item Move box, Jumping Ball: $\epsilon=0.1, \alpha=0.01$
    \item Reach: $\epsilon=0.001, \alpha=0.0001$
\end{itemize}

For all the experiments, we use batch size of $512$ and we split trajectories into chunks of size $10$. For multi-task experiments, on Figure~\ref{fig:multi_task_training}, we report 3 random seeds for each model with the best hyperparameters. Shading under the curves corresponds to 95\% confidence interval within these evaluations. We split the data on the X-axis by chunks $200000$ timesteps and the reward in these chunks is averaged. Then, we apply the rolling window smoothing with a window size of $200$.

\subsection{Adaptation experiment}

For the adaptation experiment, we train the Shared Q + DISTRAL architecture on each of the tasks. We found that the same combination of learning rate of $5e-4$ and of KL-cost of $0.01$ worked the best, so we use the same values for pre-training for all the tasks. We run 3 random seeds of pre-training and take the best performing seed to use for adaptation, therefore producing behaviour prior $\pi_{0}$ and shared features $\psi$. Then, for each task, we consider a small validation set consisting of 3 tasks which we use to choose the best adaptation hyperparameters. As for adaptation hyperparameter ranges, we consider only:
\begin{itemize}
    \item Initial inverse temperature $\alpha$: $100, 10, 1, 1e-1, 1e-2, 1e-3, 1e-4$
    \item KL-cost (inverse temperature) for DISTRAL baseline $\alpha$: $1, 0.1, 0.01, 0.001, 0.0001, 0.00001, 0.0$
\end{itemize}

For all the adaptation experiments we use learning rate of $5e-4$ and epsilon of $30$.

The best adaptation hyperparameters for IW and shared Q + IW:
\begin{itemize}
    \item Sparse 2d navigation: $\alpha=1.$
    \item Half-cheetah: $\alpha=0.01$
    \item Go to Target, Humanoid: $\alpha=1.$
    \item Go to Ring, Qudruped: $\alpha=0.1$
    \item Move box, Jumping Ball: $\alpha=0.1$
    \item Reach: $\alpha=0.1$
\end{itemize}

The best adaptation hyperparameters for DISTRAL and DISTRAL + Shared Q:
\begin{itemize}
    \item Sparse 2d navigation: $\alpha=0.1$
    \item Half-cheetah: $\alpha=0.1$
    \item Go to Target, Humanoid: $\alpha=0.1$
    \item Go to Ring, Qudruped: $\alpha=0.1$
    \item Move box, Jumping Ball: $\alpha=0.1$
    \item Reach: $\alpha=0.1$
\end{itemize}

As for baselines, $RL^2$~\citep{wang2016learning} and PEARL~\citep{rakelly2019efficient}, we use a learning rate of $5e-4$ and for PEARL we optimize a bottleneck cost from a range $10., 1., 0.1, 0.01, 0.001, 0.0001$. We use bottleneck layer dimension of $5$. The bottleneck costs per tasks are given here:
\begin{itemize}
    \item Sparse 2d navigation: $0.001$
    \item Half-cheetah: $0.001$
    \item Go to Target, Humanoid: $0.1$
    \item Go to Ring, Qudruped: $0.0001$
    \item Move box, Jumping Ball:  $1.$
    \item Reach: $0.01$
\end{itemize}

\paragraph{Adaptation protocol}
We use a fixed protocol for adaptation on all the tasks for gradient-based methods. After each unroll of sub-trajectory of size $10$, we apply 1 gradient update to the adapted parameters and after each episode we apply $50$ gradient updates. The gradient updates performed by sampling trajectories from a local replay buffer with batch size of $128$. Furthermore, for each task we act according to the behaviour prior (where appropriate) for a few exploration episodes.

\begin{itemize}
    \item Sparse 2d navigation: 5 episodes.
    \item Half-cheetah: 2 episodes.
    \item Go to Target, Humanoid: 20 episodes.
    \item Go to Ring, Qudruped: 5 episodes.
    \item Move box, Jumping Ball: 5 episodes.
    \item Reach: 5 episodes.
\end{itemize}

\begin{figure}[t!]
    \centering
   \begin{subfigure}[b]{0.3\textwidth}
         \centering
         \includegraphics[width=\textwidth]{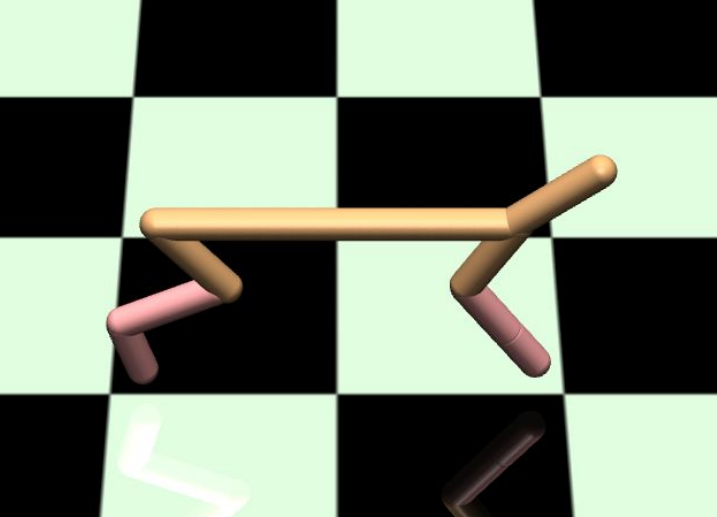}
         \caption{\emph{Half-cheetah}} 
     \end{subfigure}
    \begin{subfigure}[b]{0.3\textwidth}
         \centering
         \includegraphics[width=\textwidth]{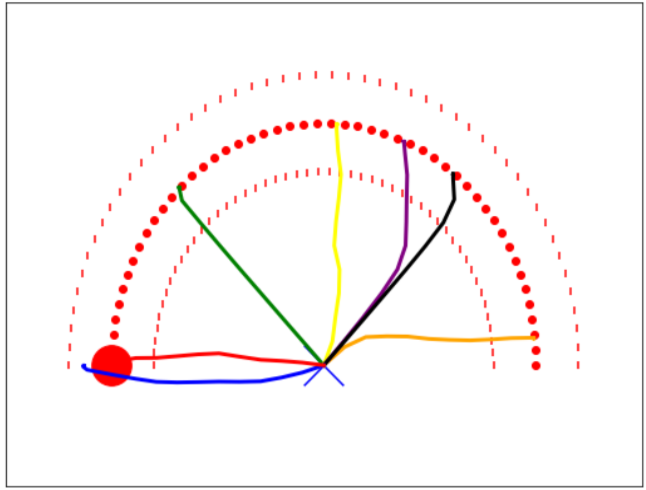}
    \caption{\emph{Sparse 2D Navigation}} 
     \end{subfigure}
     \\
     \begin{subfigure}[b]{0.3\textwidth}
         \centering
         \includegraphics[width=\textwidth]{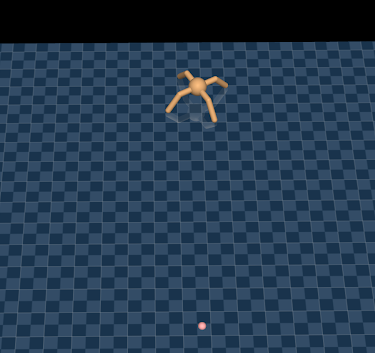}
    \caption{\emph{Go To Ring} (Ant)} 
     \end{subfigure}
     \begin{subfigure}[b]{0.3\textwidth}
         \centering
         \includegraphics[width=\textwidth]{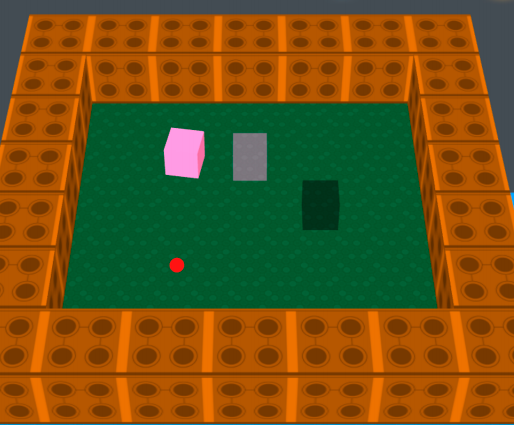}
    \caption{\emph{Move Box (Jumping ball)}} 
     \end{subfigure}
     \\
     \begin{subfigure}[b]{0.3\textwidth}
         \centering
         \includegraphics[width=\textwidth]{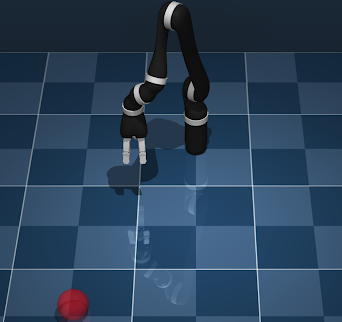}
    \caption{\emph{Reach}} 
     \end{subfigure}
     \begin{subfigure}[b]{0.3\textwidth}
         \centering
         \includegraphics[width=\textwidth]{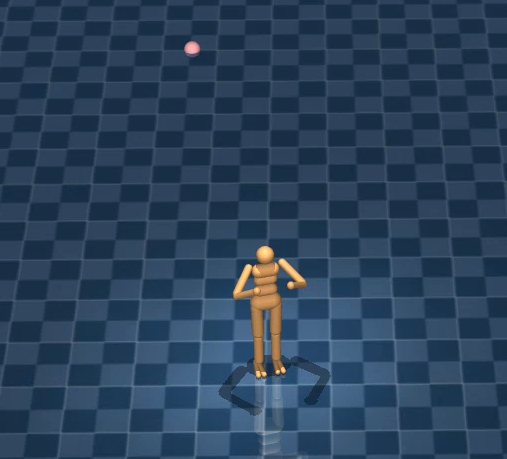}
    \caption{\emph{GTT} (Humanoid)}
     \end{subfigure}
    \caption{Tasks visualization.} 
    \label{fig:task-pictures}
\end{figure}
Curves from Figures~\ref{fig:meta_testing_theirs} and \ref{fig:meta_testing_our} plot average episodic return during adaptation, averaged over $30$ test tasks with 3 independent runs each (seeds). For each task and seed, we estimate average episodic return by averaging over the last 3 episodes.
Shading under the curves corresponds to 95\% confidence interval within these evaluations.
Results on \textit{Sparse 2D navigation} shown in Figure~\ref{fig:meta_testing_theirs} are smoothed using a rolling window of 5. No smoothing is applied for \textit{Half-cheetah velocity}.
For Figure~\ref{fig:meta_testing_our} we use a rolling window of 30.

\section{Environment Details}
\label{sec:envs}

On \emph{Go To Ring}, the agent receives a reward of 10 on achieving the target and is given an \textit{immobility penalty} of -0.005 for each time step. The episode is terminated either by achieving a target or after 10 seconds (with 20 steps per second). The task distribution is defined by $\alpha \in [0, 2 \pi]$ and $r \in [3, 7]$ which are sampled uniformly at each meta episode.  At training time, we provide only task id as task-specific information. The walker is randomly spawn at each episode in the rectangle from $[-8, 8]$. The number of training tasks is $100$, number of test tasks is $30$. We provide proprioception, global position and orientation for both behaviour prior and the agent, whereas the task identifier is provided only to the agent at training time.

For \emph{Reach}, we use a simulated Jaco robot which has to achieve a target specified in a cube with size of 0.4. Once the Jaco is within the radius of 0.05 of the target, it receives a reward of 1. The episode is terminated after 10 seconds (with 25 steps per second). At training time, we provide only task id as task-specific information. Number of training tasks is $300$, number of test tasks is $30$. We provide proprioception, global position and orientation for both behaviour prior and the agent, whereas the task identifier is provided only to the agent at training time.

For \emph{Move Box}, the reward of 10 is only given once the box is on the target. The episode is terminated either after putting the box on a target or after 20 seconds (20 steps per second). The task distribution is defined by a tuple of box and target positions, which are kept fixed for the entire meta episode. These positions are sampled uniformly in the room of size 8x8 and on maximum relative distance of 2. At training time, we provide global target position as task information. Number of training tasks is $100$, number of test tasks is $30$. We provide proprioception, global position and orientation for both behaviour prior and the agent, whereas the global target position is provided only to the agent at training time.

For \emph{GTT}, the agent receives the reward of 1.0 on achieving the target and is given an \textit{immobility penalty} of -0.005 for each time step and a penalty of -1.0 if the agent (humanoid) touches the floor with the upper body or knees. The episode is terminated either by achieving a target or after 10 seconds (with 20 steps per second). The task distribution is defined by a target position sampled uniformly on the rectangle of size 8x8. At training time, we provide only task id as task-specific information. At training time, the walker position is randomly initialized in the room at each episode, whereas for the test time, the walker initial position is kept fixed for the entire meta-episode. Number of training tasks is 100, number of test tasks is 30.  We provide proprioception, global position and orientation for both behaviour prior and the agent, whereas the task identifier is provided only to the agent at training time.

\section{Additional Results}
\label{sec:appendix_task_contex}

\begin{figure}[t!]
    \centering
    \includegraphics[width=0.9\linewidth]{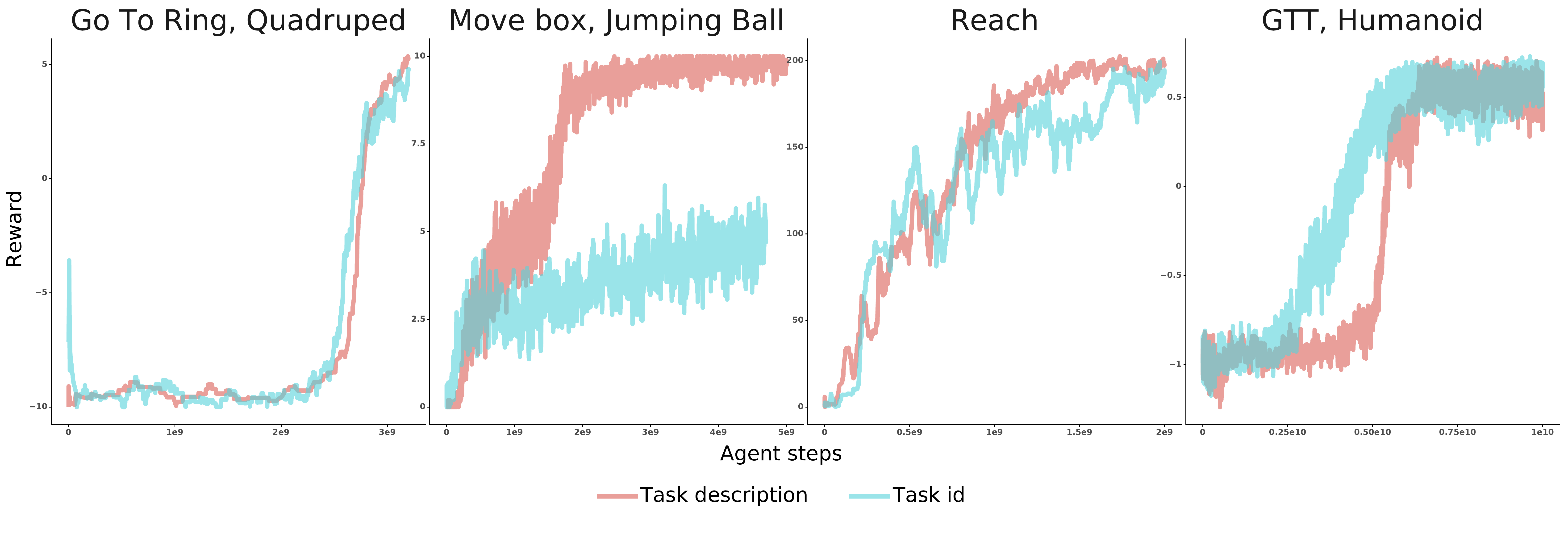}
    \caption{Meta-training performance of Shared Q method with types of task specification available at meta-training.} 
    \label{fig:meta_training_distributed}
\end{figure}

In Section~\ref{sec:transfer} ``Value Transfer'', we describe how IWPA can make use of privileged information during meta-training by mapping features $\psi$ to task specific Q-values $Q_i$, via an inner product with task features $f(g_i; w)$. Figure~\ref{fig:meta_training_distributed} reports meta-training performance of ``Shared Q'' with either $Q_i(s, a; \Phi_i) = \psi(s,a; \phi)^T w_i$ (referred as Task id) or $Q_i(s, a, g_i; \Phi_i) = \psi(s,a; \phi)^\top f(g_i;w)$ (referred as Task description), where $g_i$ is a structured task descriptor. The latter yields a qualitative difference on Move Box, where this information represents a global position of a target location. This confirms that using rich privileged information during meta-training, is important to scale meta and transfer learning approaches to more challenging domains.

\section{Ablations}
\label{sec:appendix_ablation}

\begin{figure}[t!]
    \centering
    \includegraphics[width=0.9\linewidth]{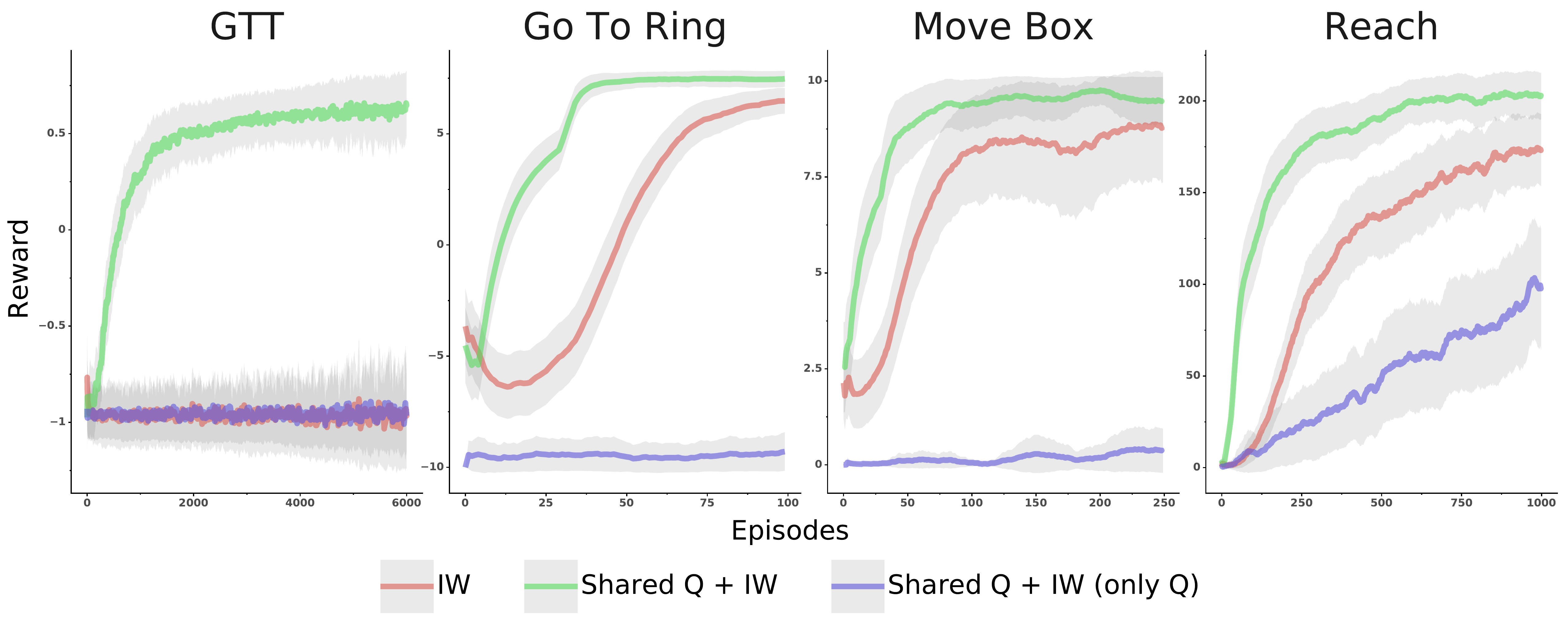}
    \caption{Ablation demonstrating "Shared Q + IW" architecture where both value function and the prior policy are reloaded. In "Shared Q + IW", only the Q-function is reloaded, and in "IW", only the behaviour prior is reloaded.} 
    \label{fig:shared_q_ablation}
\end{figure}

The method IWPA described in Section~\ref{sec:transfer} and in Algorithm~\ref{alg:iwpa} relies on both behaviour prior $\pi_{0}$ and learnt Q-function features $\psi$. Furthermore, based on the transfer learning results presented in Figure~\ref{fig:meta_testing_our}, it may seem that state-action value function features are a crucial component for the transfer. In this section, we provide an ablation, where we show that without a behaviour prior, these features only do not transfer. Therefore, the combination of both, behaviour prior and value features is important. The results are given in Figure~\ref{fig:shared_q_ablation}. As we can see, the architecture which uses both components, "Shared Q + IW" works very well, whereas the one which reloads only the value features fails to learn.

\end{document}